\documentclass[10pt,twocolumn,letterpaper]{article}

\usepackage{cvpr}
\usepackage{times}
\usepackage{epsfig}
\usepackage{graphicx}
\usepackage{amsmath}
\usepackage{amssymb}
\usepackage{diagbox}
\usepackage{multirow}
\usepackage[boxed]{algorithm2e}
\usepackage{siunitx}
\usepackage{gensymb}
\usepackage{pifont}
\usepackage{xcolor}

\newcommand{\cmark}{\ding{51}}
\newcommand{\xmark}{\ding{55}}

\usepackage[pagebackref=true,breaklinks=true,letterpaper=true,colorlinks,bookmarks=false]{hyperref}

\cvprfinalcopy 

\def\cvprPaperID{1381} 

\begin{document}

\title{CRAVES: Controlling Robotic Arm with a Vision-based Economic System}

\author{Yiming Zuo\textsuperscript{1}\thanks{This work was done when the first author was an intern at the Johns Hopkins University. The first two authors contributed equally.}, Weichao Qiu\textsuperscript{2}\footnotemark[1], Lingxi Xie\textsuperscript{2,4}, Fangwei Zhong\textsuperscript{3}, Yizhou Wang\textsuperscript{3,5,6}, Alan L. Yuille\textsuperscript{2}\\
\textsuperscript{1}Tsinghua University\quad\quad\textsuperscript{2}Johns Hopkins University\quad\quad\textsuperscript{3}Peking University\\
\textsuperscript{4}Noah's Ark Lab, Huawei Inc.\quad\textsuperscript{5}Peng Cheng Laboratory\quad\textsuperscript{6}DeepWise AI Lab\\
{\tt\small \{zuoyiming17,qiuwch,198808xc,zfw1226,alan.l.yuille\}@gmail.com,}\quad{\tt\small yizhou.wang@pku.edu.cn}}
\maketitle

\begin{abstract}
Training a robotic arm to accomplish real-world tasks has been attracting increasing attention in both academia and industry. This work discusses the role of computer vision algorithms in this field. We focus on low-cost arms on which no sensors are equipped and thus all decisions are made upon visual recognition, e.g., real-time 3D pose estimation. This requires annotating a lot of training data, which is not only time-consuming but also laborious.

In this paper, we present an alternative solution, which uses a 3D model to create a large number of synthetic data, trains a vision model in this virtual domain, and applies it to real-world images after domain adaptation. To this end, we design a semi-supervised approach, which fully leverages the geometric constraints among keypoints. We apply an iterative algorithm for optimization. Without any annotations on real images, our algorithm generalizes well and produces satisfying results on 3D pose estimation, which is evaluated on two real-world datasets. We also construct a vision-based control system for task accomplishment, for which we train a reinforcement learning agent in a virtual environment and apply it to the real-world. Moreover, our approach, with merely a 3D model being required, has the potential to generalize to other types of multi-rigid-body dynamic systems.

\end{abstract}

\begin{figure*}[!th]
\centering
\includegraphics[width=0.9\textwidth]{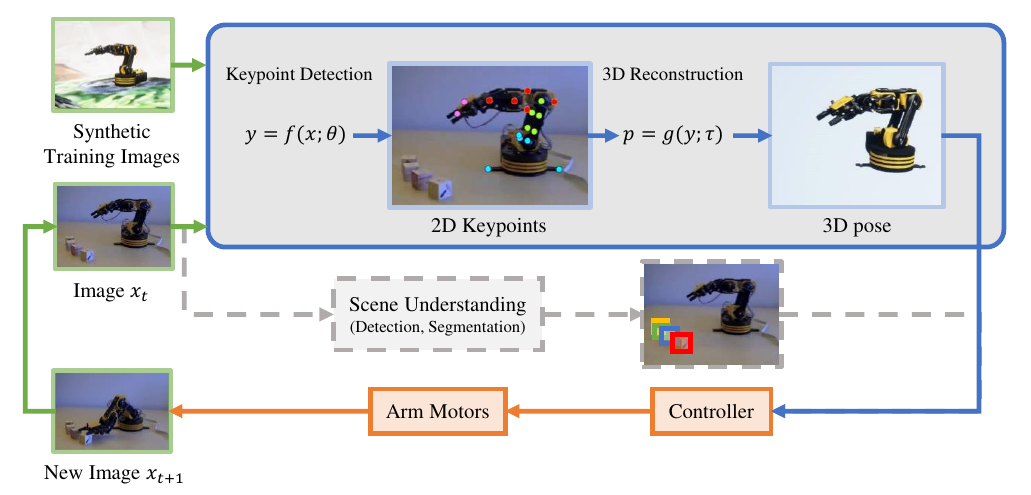}
\caption{An overview of our system (best viewed in color). The goal is to accomplish tasks using camera as the only sensor. The vision module detects 2D keypoints of the arm and then computes its 3D pose. The control module uses the 3D pose estimation to determine the next move and feeds it to the motors. In our setup, scene understanding is achieved by directly providing the 3D location of targets.}
\label{Fig:Overview}
\vspace{-0.2cm}
\end{figure*}

\section{Introduction}
\label{Introduction}

Precise and agile robotic arms have been widely used in the assembly industry for decades, but the adaptation of robots to domestic use is still a challenging topic. This task can be made much easier if vision input are provided and well utilized by the robots. A typical example lies in autonomous driving~\cite{geiger2013vision}. In the area of robotics, researchers have paid more and more attentions to vision-based robots and collected large-scale datasets, {\em e.g.}, for object grasping~\cite{levine2018learning}\cite{pinto2016supersizing} and block stacking~\cite{hundt2018training}. However, the high cost of configuring a robotic system largely limits researchers from accessing these interesting topics.

This work aims at equipping a robotic system with computer vision algorithms, {\em e.g.}, predicting its real-time status using an external camera, so that researchers can control them with a high flexibility, {\em e.g.}, mimicking the behavior of human operators. In particular, we build our platform upon a low-cost robotic arm named OWI-535 which can be purchased from Amazon\footnote{{\sf https://www.amazon.com/dp/B0017OFRCY/}} for less than $\$40$. The downside is that this arm has no sensors and thus it totally relies on vision inputs\footnote{Even when the initialized status of the arm is provided and each action is recorded, we cannot accurately compute its real-time status because each order is executed with large variation -- even the battery level can affect.} -- on the other hand, we can expect vision inputs to provide complementary information in sensor-equipped robotic systems. We chose this arm for two reasons. (i) {\bf Accessibility}: the cheap price reduces experimental budgets and makes our results easy to be reproduced by lab researchers (poor vision people :( ). (ii) {\bf Popularity}: users around the world uploaded videos to YouTube recording how this arm was manually controlled to complete various tasks, {\em e.g.}, picking up tools, stacking up dices, {\em etc}. These videos were captured under substantial environmental changes including viewpoint, lighting condition, occlusion and blur. This raises real-world challenges which are very different from research done in a lab environment.

Hence, the major technical challenge is to {\bf train a vision algorithm to estimate the 3D pose of the robotic arm}. Mathematically, given an input image $\mathbf{x}$, a vision model ${\mathbb{M}}:{\mathbf{p}}={\mathbf{h}\!\left(\mathbf{x};\boldsymbol{\theta}\right)}$ is used to predict $\mathbf{p}$, the real-time 3D pose of the arm, where $\boldsymbol{\theta}$ denotes the learnable parameters, {\em e.g.}, in the context of deep learning~\cite{lecun2015deep}, network weights. Training such a vision model often requires a considerable amount of labeled data. One option is to collect a large number of images under different environments and annotate them using crowd-sourcing, but we note a significant limitation as these efforts, which take hundreds of hours, are often not transferable from one robotic system to another. In this paper, we suggest an alternative solution which borrows a 3D model and synthesizes an arbitrary amount of labeled data in the virtual world with almost no cost, and later adapts the vision model trained on these virtual data to real-world scenarios.

This falls into the research area of domain adaptation~\cite{csurka2017domain}. Specifically, the goal is to train $\mathbb{M}$ on a virtual distribution ${\mathbf{x}^\mathrm{V}}\sim{\mathcal{P}^\mathrm{V}}$ and then generalize it to the real distribution ${\mathbf{x}^\mathrm{R}}\sim{\mathcal{P}^\mathrm{R}}$. We achieve this goal by making full use of a strong property, that the spatial relationship between keypoints, {\em e.g.}, the length of each bone, is fixed and known. This is to say, although the target distribution $\mathcal{P}^\mathrm{R}$ is different from $\mathcal{P}^\mathrm{V}$ and data in $\mathcal{P}^\mathrm{R}$ remain unlabeled, the predicted keypoints should strictly obey some geometric constraints $\boldsymbol{\tau}$. To formulate this, we decompose $\mathbb{M}$ into two components, namely ${\mathbb{M}_1}:{\mathbf{y}}={\mathbf{f}\!\left(\mathbf{x};\boldsymbol{\theta}\right)}$ for keypoint detection and ${\mathbb{M}_2}:{\mathbf{p}}={\mathbf{g}\!\left(\mathbf{y};\boldsymbol{\tau}\right)}$ for 3D pose estimation, respectively. Here, $\mathbb{M}_2$ is parameter-free and thus cannot be optimized, so we train $\mathbb{M}_1$ on $\mathcal{P}^\mathrm{V}$ and hope to adapt it to $\mathcal{P}^\mathrm{R}$, and $\mathbf{y}$ becomes a hidden variable. We apply an iterative algorithm to infer ${\mathbf{p}^\star}={\arg\max_{\mathbf{p}}\int\mathrm{Pr}\!\left(\mathbf{y};\boldsymbol{\theta}\mid\mathbf{x}\right)\cdot\mathrm{Pr}\!\left(\mathbf{p};\boldsymbol{\tau}\mid\mathbf{y}\right)\mathrm{d}\mathbf{y}}$, and the optimal $\mathbf{y}^\star$ determined by $\mathbf{p}^\star$ serves as the guessed label, which is used to fine-tune $\mathbb{M}_1$. Eventually, prediction is achieved without any annotations in the target domain.

We design two benchmarks to evaluate our system. The first one measures pose estimation accuracy, for which we manually annotate two image datasets captured in our lab and crawled from YouTube, respectively. Our algorithm, trained on labeled virtual data and fine-tuned with unlabeled lab data, achieves a mean angular error of $4.81^\circ$, averaged over $4$ joints. This lays the foundation of the second benchmark in which we create an environment for the arm to accomplish a real-world task, {\em e.g.}, touching a specified point. Both quantitative (in distance error and success rates) and qualitative (demos are provided in the supplementary material) results are reported. Equipped with reinforcement learning, our vision-based algorithm achieves comparable accuracy with human operators. All our data and code have been released at our website, \url{https://qiuwch.github.io/craves.ai}.

In summary, the contribution of this paper is three-fold. First, we design a complete framework to achieve satisfying accuracy in task accomplishment with a low-cost, sensor-free robotic arm. Second, we propose a vision algorithm involving training in virtual environment and domain adaptation, and verify its effectiveness in a typical multi-rigid-body system. Third, we develop a platform with two real-world datasets and a virtual environment so as to facilitate future research in this field.


\section{Related Work}
\label{RelatedWork}

\noindent
$\bullet$\quad{\bf Vision-based Robotic Control}

Vision-based robotic control is attracting more and more attentions. Compared with conventional system relying on specific sensors, \eg IMU and rotary encoder, vision has the flexibility to adapt to complex and novel tasks. Recent progress of computer vision makes vision-based robotic control more feasible. Besides using vision algorithms as a perception module, researchers are also exploring training an end-to-end control system purely from vision~\cite{jang2017end}\cite{levine2016end}\cite{luo2018end}. To this end, researchers collected large datasets for various tasks, including grasping~\cite{levine2018learning}\cite{pinto2016supersizing}, block stacking~\cite{hundt2018training}, autonomous driving~\cite{geiger2013vision}\cite{yu2018bdd100k}, {\em etc}.

On the other hand, training a system for real-world control tasks is always time-consuming, and high-accuracy sensor-based robots are expensive , both of which have prevented a lot of vision researchers from entering this research field. For the first issue, people turned to use simulators such as MuJoCo~\cite{todorov2012mujoco} and Gazebo~\cite{koenig2004design} so as to accelerate training processes, {\em e.g.}, with reinforcement learning, and applied to real robots, {\em e.g.}, PR2~\cite{tzeng2015adapting}, Jaco~\cite{rusu2016sim} and KUKA IIWA~\cite{bousmalis2018using}(ranging from $\$50,000$  to $\$200,000$). For the second issue, although low-cost objects ({\em e.g.}, toy cars~\cite{kahn2017uncertainty}) have been used to simulate real-world scenarios, low-cost robotic arms were rarely used, mainly due to the limitation caused by the imprecise motors and/or sensors, so that conventional control algorithms are difficult to be applied. For instance, Lynxmotion Arm is an inexpensive ($\$300$) robotic arm used for training reinforcement learning algorithms~\cite{rahmatizadeh2018vision}\cite{deisenroth2012learning}. The control of this arm was done using a hybrid of camera and servo-motor, which provides joint angle. This paper uses an even cheaper ($\$40$) and more popular robotic arm named OWI-535, which merely relies on vision inputs from an external camera. To the best of our knowledge, this arm has never been used for automatic task accomplishment, because lacking of sensors.

\begin{figure}
\centering
\includegraphics[width=0.8\linewidth]{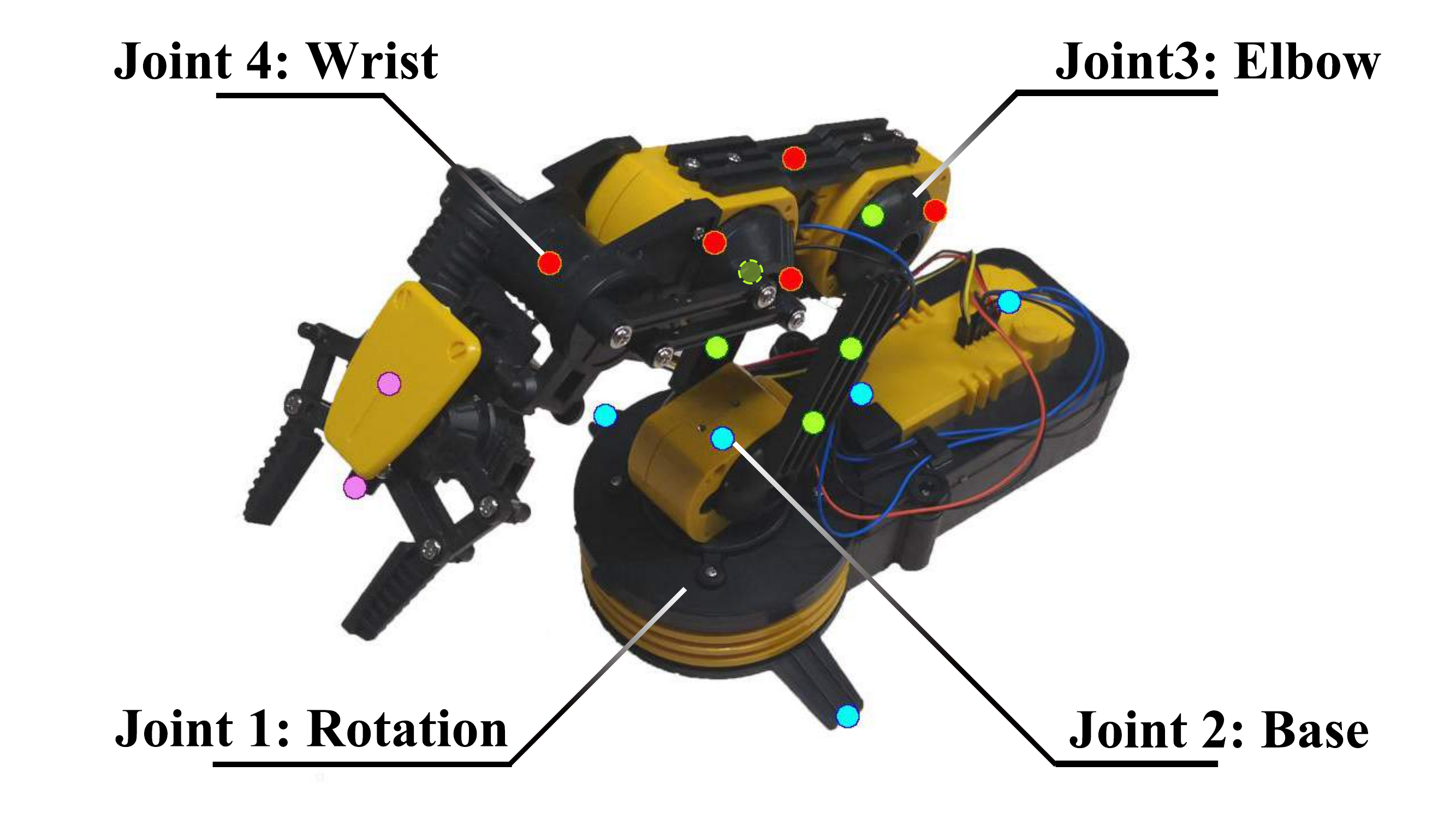}
\caption{Here shows the $4$ joints and $17$ keypoints of OWI-535 used in our experiment. Each joint is assigned a specific name. Color of keypoint correspond to the part to which it belongs.}
\label{Fig:KeypointDefinition}
\vspace{-0.2cm}
\end{figure}

\vspace{0.2cm}
\noindent
$\bullet$\quad{\bf Computer Vision with Synthetic Data}

Synthetic data have been widely applied to many computer vision problems in which annotations are difficult to obtain, such as optical flow~\cite{butler2012naturalistic}, object tracking~\cite{gaidon2016virtual}\cite{luo2018journal}\cite{zhong2018advat}, human parsing~\cite{varol2017learning}, VQA~\cite{johnson2017clevr}, $6$-D pose estimation~\cite{su2015render}\cite{sundermeyer2018implicit}, semantic segmentation~\cite{hoffman2016fcns}, {\em etc}.

Domain adaptation is an important stage to transfer models trained on synthetic data to real scenarios. There are three major approaches, namely, domain randomization~\cite{tobin2017domain}\cite{tremblay2018training}, adversarial training~\cite{hoffman2018cycada}\cite{mahmood2018unsupervised}\cite{shrivastava2017learning}\cite{tzeng2017adversarial} and joint supervision~\cite{li2017deep}. A more comprehensive survey on domain adaptation is available in~\cite{csurka2017domain}. As an alternative solution, researchers introduced intermediate representation ({\em e.g.}, semantic segmentation) to bridge the domain gap~\cite{hong2018virtual}. In this work, we focus on semi-supervised learning with the assistance of domain randomization. The former method is mainly based on 3D priors obtained from modeling the geometry of the target object~\cite{gao2017exploiting}\cite{pavlakos2018learning}. Previously, researchers applied parameterized 3D models to refine the parsing results of humans~\cite{bogo2016keep}\cite{omran2018neural} or animals~\cite{zuffi2018lions}, or fine-tune the model itself~\cite{pavlakos2018learning}. The geometry of a robotic system often has a lower degree of freedom, which enables strong shape constraints to be used for both purposes, {\em i.e.}, prediction refinement and model fine-tuning.

\section{Approach}
\label{Approach}

\subsection{System Overview}
\label{Approach:Overview}

We aim at designing a vision-based system to control a sensor-free robotic arm to accomplish real-world tasks. Our system, as illustrated in Figure~\ref{Fig:Overview}, consists of three major components working in a loop. The first component, {\em data acquisition}, samples synthetic images $\mathbf{x}$ from a virtual environment for training and real images from an external camera for real-time control. The second component is {\em pose estimation}, an algorithm ${\mathbf{p}}={\mathbf{h}\!\left(\mathbf{x};\boldsymbol{\theta}\right)}$ which produces the 3D pose (joint angles) of the robotic arm (see Figure~\ref{Fig:KeypointDefinition} for the definition of four joints). The third component is a {\em controller}, which takes $\mathbf{p}$ as input, determines an action for the robotic arm to take, and therefore triggers a new loop.

Note that data acquisition (Section~\ref{Approach:Data}) may happen in both virtual and real environments -- our idea is to collect cheap training data in the virtual domain, train a vision model and tune it into one that works well in real world. The core of this paper is pose estimation (Section~\ref{Approach:PoseEstimation}), which is itself an important problem in computer vision, and we investigate it from the perspective of domain transfer. While studying motion control (Section~\ref{Approach:MotionPlanning}) is also interesting yet challenging, it goes out of the scope of this paper, so we setup a relatively simple environment and apply an reinforcement learning algorithm.

\subsection{Data Acquisition}
\label{Approach:Data}

The OWI-535 robotic arm has $5$ motors named {\em rotation}, {\em base}, {\em elbow}, {\em wrist} and {\em gripper}. Among them, the status of the {\em gripper} is not necessary for motion planing and thus it is simply ignored in this paper. The range of motion for the first 4 motors are $270\degree$, $180\degree$, $300\degree$ and $120\degree$, respectively. 

In order to collect training data with low costs, we turn to the virtual world. We download a CAD model of the arm with exactly the same appearance as the real one which was constructed using Unreal Engine 4 (UE4)\footnote{{\sf https://3dwarehouse.sketchup.com/model/u21290a3e-f8ef-46da-985c-9aa56b0dee53/Maplin-OWI-ROBOTIC-ARM}}. Using Maya, we implement its motion system which was not equipped in the original model. The angle limitation as well as the collision boundary of each joint is also manually configured. The CAD model of OWI-535 has $170\rm{,}648$ vertices in total, among which, we manually annotate $17$ visually distinguishable vertices as our keypoints, as shown in Figure~\ref{Fig:KeypointDefinition}. This number is larger than the degree-of-freedom of the system ($6$ camera parameters and $4$ joint angles), meanwhile reasonably small so that all keypoints can be annotated on real images for evaluation. The images and annotations are collected from UE4 via UnrealCV~\cite{qiu2017unrealcv}.

We create real-world dataset from two sources for benchmark and replication purpose. The first part of data are collected from an arm in our own lab, and we maximally guarantee the variability in its pose, viewpoint and background. The second part of data are crawled from YouTube, on which many users uploaded videos of playing with this arm. Both subsets raise new challenges which are not covered by virtual data, such as motion blur and occlusion, to our vision algorithm, with the second subset being more challenging as the camera intrinsic parameters are unknown and the arm may be modded for various purposes. We manually annotate the $17$ keypoints on these images, which typically takes one minute for a single frame.

More details of these datasets are covered in Section~\ref{Experiments}.

\subsection{Transferable 3D Pose Estimation}
\label{Approach:PoseEstimation}

We first define some terminologies used in this paper. Let $\mathbf{p}$ define all parameters that determine the arm's position in the image. In our implementation, $\mathbf{p}$ has $10$ dimensions: $6$ camera extrinsic parameters (location, rotation) and $4$ angles of motors.
${\mathbf{y}}\in{\mathbb{R}^{17\times2}}$ and ${\mathbf{z}}\in{\mathbb{R}^{17\times3}}$ are the locations of keypoints in 2D and 3D, respectively. Both $\mathbf{y}$ and $\mathbf{z}$ are {\em deterministic} functions of $\mathbf{p}$.

The goal is to design a function ${\mathbf{p}}={\mathbf{h}\!\left(\mathbf{x};\boldsymbol{\theta}\right)}$ which receives an image $\mathbf{x}$ and outputs the pose vector $\mathbf{p}$ which defines the pose of the object. $\boldsymbol{\theta}$ denotes the learnable parameters, {\em e.g.}, network weights in the context of deep learning. The keypoints follow the same geometry constraints in both virtual and real domains. In order to fully utilize these constraints, we decompose $\mathbf{h}\!\left(\cdot\right)$ into two components, namely, 2D keypoint detection ${\mathbb{M}_1}:{\mathbf{y}}={\mathbf{f}\!\left(\mathbf{x};\boldsymbol{\theta}\right)}$ and 3D pose estimation ${\mathbb{M}_2}:{\left(\mathbf{p},\mathbf{z}\right)}={\mathbf{g}\!\left(\mathbf{y};\boldsymbol{\tau}\right)}$. Here, $\boldsymbol{\tau}$ is a fixed set of equations corresponding to the geometric constraints, {\em e.g.}, the length between two joints. This is to say, $\mathbb{M}_1$ is trained to optimize $\boldsymbol{\theta}$ while $\mathbb{M}_2$ is a parameter-free algorithm which involves fitting a few fixed arithmetic equations.

To alleviate the expense of data annotation, we apply a setting known as semi-supervised learning~\cite{zhu2006semi} which contains two parts of training data. First, a labeled set of training data ${\mathcal{D}_1}={\left\{\left(\mathbf{x}_n,\mathbf{y}_n\right)\right\}_{n=1}^N}$ is collected from the virtual environment. This process is performed automatically with little cost, and also easily transplanted to other robotic systems with a 3D model available. Second, an unlabeled set of image data ${\mathcal{D}_2}={\left\{\tilde{\mathbf{x}}_m\right\}_{m=1}^M}$ is provided, while the corresponding label $\tilde{\mathbf{y}}_m$ for each $\tilde{\mathbf{x}}_m$ remains unknown. We use $\mathcal{P}^\mathrm{V}$ and $\mathcal{P}^\mathrm{R}$ to denote the virtual and real image distributions, {\em i.e.}, ${\mathbf{x}_n}\sim{\mathcal{P}^\mathrm{V}}$ and ${\tilde{\mathbf{x}}_m}\sim{\mathcal{P}^\mathrm{R}}$, respectively. Since $\mathcal{P}^\mathrm{V}$ and $\mathcal{P}^\mathrm{R}$ can be different in many aspects, we cannot expect a model trained on $\mathcal{D}_1$ to generalize sufficiently well on $\mathcal{D}_2$.

The key is to bridge the gap between $\mathcal{P}^\mathrm{V}$ and $\mathcal{P}^\mathrm{R}$. One existing solution works in an {\em explicit} manner, which trains a mapping $\mathbf{r}\!\left(\cdot\right)$, so that when we sample $\tilde{\mathbf{x}}_m$ from $\mathcal{P}^\mathrm{R}$, $\mathbf{r}\!\left(\tilde{\mathbf{x}}_m\right)$ maximally mimics the distribution of $\mathcal{P}^\mathrm{V}$. This is achieved by unpaired image-to-image translation~\cite{zhu2017unpaired}, which was verified effective in some vision tasks~\cite{hoffman2018cycada}. However, in our problem, an additional cue emerges, claiming that the source and target images have the same label distribution, {\em i.e.}, both scenarios aim at estimating the pose of exactly the same object, so we can make use of this cue to achieve domain adaptation in an {\em implicit} manner. In practice, we do semi-supervised training by providing the system with unlabeled data. Our approach exhibits superior transfer ability in this specific task, while we preserve the possibility of combining both manners towards higher accuracy.

To this end, we reformulate $\mathbb{M}_1$ and $\mathbb{M}_2$ in a probabilistic style. $\mathbb{M}_1$ produces a distribution ${\mathcal{F}\!\left(\mathbf{x};\boldsymbol{\theta}\right)}\ni{\mathbf{y}}$, and similarly, $\mathbb{M}_2$ outputs ${\mathcal{G}\!\left(\mathbf{y};\boldsymbol{\tau}\right)}\ni{\left(\mathbf{p},\mathbf{z}\right)}$. Here, the goal is to maximize the marginal likelihood of $\left(\mathbf{p},\mathbf{z}\right)$ while $\mathbf{y}$ remains a latent variable:
\begin{equation}
\label{Eqn:Optimization}
{\left(\mathbf{p}^\star,\mathbf{z}^\star\right)}={\arg\max_{\left(\mathbf{p},\mathbf{z}\right)}\int\mathrm{Pr}\!\left(\mathbf{y};\boldsymbol{\theta}\mid\mathbf{x}\right)\cdot\mathrm{Pr}\!\left(\mathbf{p},\mathbf{z};\boldsymbol{\tau}\mid\mathbf{y}\right)\mathrm{d}\mathbf{y}}.
\end{equation}
There is another option, which directly computes ${\mathbf{y}^\star}={\arg\max_\mathbf{y}\mathrm{Pr}\!\left(\mathbf{y};\boldsymbol{\theta}\mid\mathbf{x}\right)}$ and then infers $\mathbf{p}^\star$ and $\mathbf{z}^\star$ from $\mathbf{y}^\star$. We do not take it because we trust $\mathrm{Pr}\!\left(\mathbf{p},\mathbf{z};\boldsymbol{\tau}\mid\mathbf{y}\right)$ more than $\mathrm{Pr}\!\left(\mathbf{y};\boldsymbol{\theta}\mid\mathbf{x}\right)$, since the former is formulated by strict geometric constraints. Eqn~\eqref{Eqn:Optimization} can be solved using an iterative algorithm, starting with a model $\mathcal{F}\!\left(\mathbf{x};\boldsymbol{\theta}\right)$ pre-trained in the virtual dataset.

In the first step, we fix $\boldsymbol{\theta}$ and infer $\mathcal{F}\!\left(\mathbf{x};\boldsymbol{\theta}\right)$. This is done by cropping the input image to $256\times256$ and feeding it to a stacked hourglass network~\cite{newell2016stacked} with $2$ stacks. The network produces ${K}={17}$ heatmaps, each of which, sized $64\times64$, corresponds to a keypoint. These heatmaps are taken as input data of $\mathcal{G}\!\left(\mathbf{y};\boldsymbol{\tau}\right)$ which estimates $\mathbf{p}$ and $\mathbf{z}$ as well as $\mathbf{y}$. This is done by making use of geometric constraints $\boldsymbol{\tau}$, which appears as a few linear equations with fixed parameters, {\em e.g.}, the length of each bone of the arm. This is a probabilistic model and we apply an iterative algorithm (see Section~\ref{Approach:Details}) to find an approximate solution $\mathbf{y}'$, $\mathbf{p}'$ and $\mathbf{z}'$. Note that $\mathbf{y}'$ is not necessarily the maximum in $\mathcal{F}\!\left(\mathbf{x};\boldsymbol{\theta}\right)$.

\begin{figure}[tb]
\centering
\includegraphics[width=\linewidth]{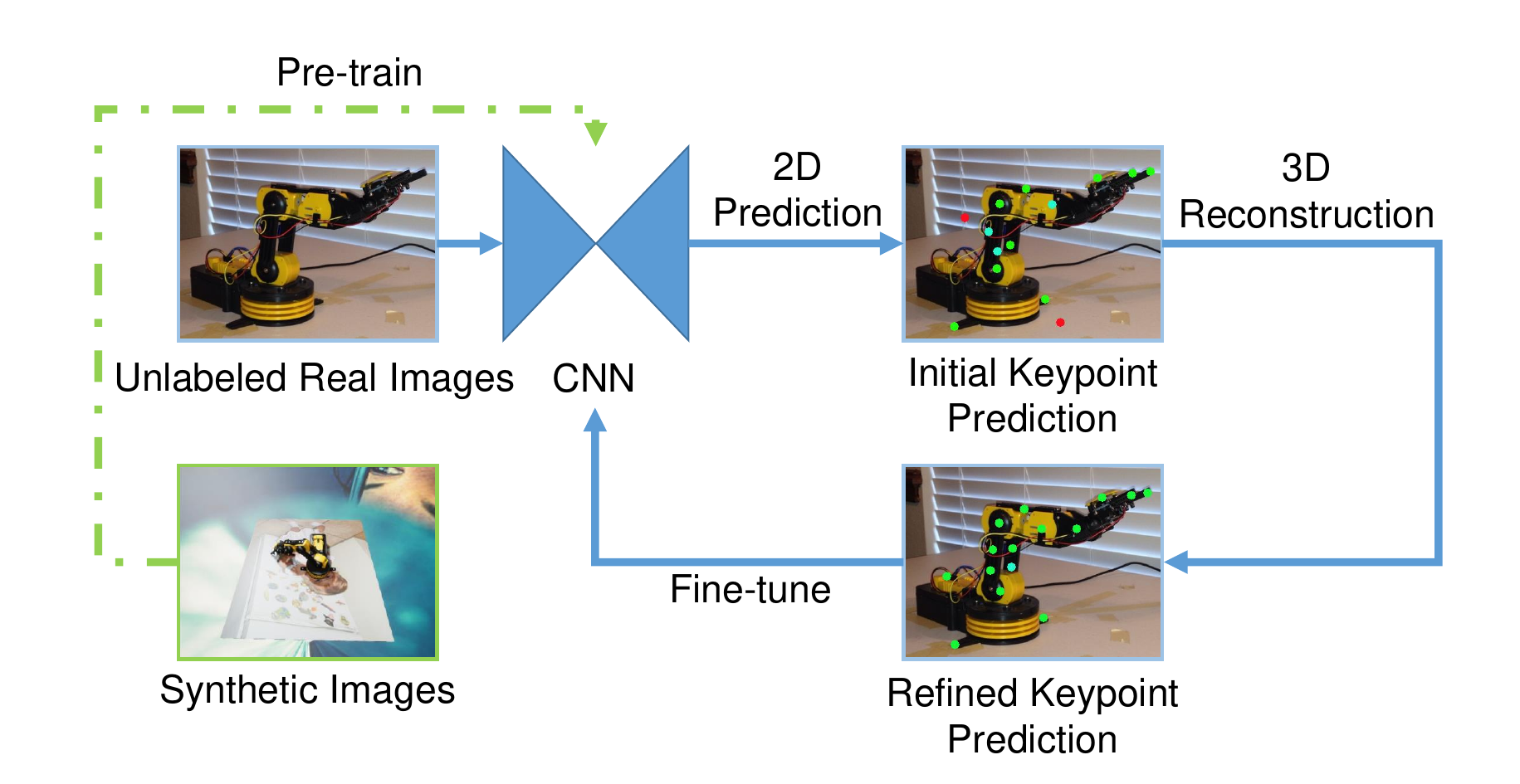}
\caption{The pipeline of transferable 3D pose estimation (best viewed in color). The initial prediction may contain both accurate (\textcolor{green}{green}) and inaccurate (\textcolor{cyan}{cyan}) keypoints, and even outliers (\textcolor{red}{red}). By introducing a 3D-prior constraint, we obtain a refined keypoint prediction, which is used to fine-tune the neural network.}
\vspace{-0.4cm}
\label{Fig:DA_Pipeline}
\end{figure}

In the second step, we take the optimal $\mathbf{y}'$ to update $\boldsymbol{\theta}$. As $\mathcal{F}\!\left(\mathbf{x};\boldsymbol{\theta}\right)$ is a deep network, this is often achieved by gradient back-propagation. We incorporate this iterative algorithm with stochastic gradient descent. In each basic unit known as an {\em epoch}, each step is executed only once. Although convergence is most often not achieved, we continue with the next epoch, which brings more informative supervision. Compared with solving Eqn~\eqref{Eqn:Optimization} directly, this strategy improves the efficiency in the training stage, {\em i.e.}, a smaller number of iterations is required. 
Figure \ref{Fig:DA_Pipeline} shows an illustration of our transferable pose estimation pipeline.

\subsection{Motion Control}
\label{Approach:MotionPlanning}

In order to control the arm to complete tasks, we need a motion control module which takes the estimated 3D pose as input and outputs an action to achieve the goal. The motion control policy $a_t=\pi(s_t, g_t)$ is learned via a deep reinforcement learning algorithm. $s_t$ is the state about the environment at time $t$, \eg the arm pose $\mathbf{p}$. $g_t$ represents the goal, \eg target location. $a_t$ is the control signal for each joint in our system. The policy is learned in our virtual environment, and optimized by Deep Deterministic Policy Gradient (DDPG)~\cite{ddpg}. Our experiment shows that using arm pose as input, the policy learned in the virtual environment can be directly applied to the real world.

\subsection{Implementation Details}
\label{Approach:Details}

\noindent
$\bullet$\quad{\bf Training Data Variability}

Our approach involves two parts of training data, namely, a virtual subset to pre-train 2D keypoint detection, and an unlabeled real subset for fine-tuning. In both parts, we change the background contents of {\em each} training image so as to facilitate data variability and thus alleviate over-fitting.

In the virtual domain, background can be freely controlled by the graphical renderer. In this scenario, we place the arm on a board, under a sky sphere, and the background of the board and the sphere are both randomly sampled from the MS-COCO dataset~\cite{lin2014microsoft}. In the real domain, however, background parsing is non-trivial yet can be inaccurate. To prevent this difficulty, we create a special subset for fine-tuning, in which all images are captured in a clean environment, {\em e.g.}, in front of a white board, which makes it easy to segment the arm with a pixel-wise color-based filter, and then place it onto a random image from the MS-COCO dataset. We observe consistent accuracy gain brought by these simple techniques.

\vspace{0.2cm}
\noindent
$\bullet$\quad{\bf Joint Keypoint Detection and Pose Estimation}

We use an approximate algorithm to find the $\mathbf{y}'$ (as well as $\mathbf{p}'$ and $\mathbf{z}'$) that maximizes $\mathrm{Pr}\!\left(\mathbf{y};\boldsymbol{\theta}\mid\mathbf{x}\right)\cdot\mathrm{Pr}\!\left(\mathbf{p},\mathbf{z};\boldsymbol{\tau}\mid\mathbf{y}\right)$ in Eqn~\eqref{Eqn:Optimization}, because an accurate optimization is mathematically intractable. We first compute ${\mathbf{y}'}={\arg\max_{\mathbf{y}}\mathcal{F}\!\left(\mathbf{x};\boldsymbol{\theta}\right)}$ which maximizes $\mathrm{Pr}\!\left(\mathbf{y};\boldsymbol{\theta}\mid\mathbf{x}\right)$. This is performed on the heatmap of each 2D keypoint individually, which produces not only the most probable $\mathbf{y}_k'$ but also a  score $c_k$ indicating its confidence. We first filter out all keypoints with a threshold $\xi$, {\em i.e.}, all keypoints with ${c_k}<{\xi}$ are considered unknown (and thus completely determined by geometric prior) in the following 3D reconstruction module. This is to maximally prevent the impact of outliers. In practice, we use ${\xi}={0.3}$ and our algorithm is not sensitive to this parameter.

Next, we recover the 3D pose using these 2D keypoints, {\em i.e.}, maximizing $\mathrm{Pr}\!\left(\mathbf{p},\mathbf{z};\boldsymbol{\tau}\mid\mathbf{y}=\mathbf{y}'\right)$. Under the assumption of perspective projection, that each keypoint ${\mathbf{y}_k}\in{\mathbb{R}^2}$ is the 2D projection of a 3D coordinate ${\mathbf{z}_k}\in{\mathbb{R}^3}$, which can be written in a linear equation:
\begin{equation}
\label{Eqn:AffineTransformation}
{\left[\mathbf{y}|\mathbf{1}\right]^\top}\cdot\hat{\mathbf{S}}=\mathbf{K}\cdot{\left[\mathbf{R}|\mathbf{T}\right]\cdot\left[\mathbf{z}|\mathbf{1}\right]^\top}.
\end{equation}
Here, ${\mathbf{y}}\in{\mathbb{R}^{K\times2}}$ and ${\mathbf{z}}\in{\mathbb{R}^{K\times3}}$ are 2D and 3D coordinate matrices, respectively, and ${\mathbf{1}}\in{\mathbb{R}^{K\times1}}$ is an all-one vector. $\mathbf{K}\in{\mathbb{R}^{3\times3}}$ is the camera intrinsic matrix, which is constant for a specific camera. ${\mathbf{S}}\in{\mathbb{R}^K}$, ${\mathbf{R}}\in{\mathbb{R}^{3\times3}}$ and ${\mathbf{T}}\in{\mathbb{R}^{3\times1}}$ denote the scaling vector, rotation matrix and translation vector, respectively, all of which are determined by $\mathbf{p}$. For each keypoint $k$, $\mathbf{z}^k$ is determined by the motor transformation ${\mathbf{z}^k}={\mathbf{z}_0^k\cdot\mathbf{W}^k}$, where ${\mathbf{z}_0}\in{\mathbb{R}^{K\times3}}$ is a constant matrix indicating the coordinates of all keypoints when motor angles are $0$, and ${\mathbf{W}^k}\in{\mathbb{R}^{3\times3}}$ is the motor transformation matrix for the $k$th keypoint, which is also determined by $\mathbf{p}$. ${\hat{\mathbf{S}}}={\mathbf{diag}\!\left(\mathbf{S}\right)}$ is the scaling matrix. Due to the inaccuracy in prediction ($\mathbf{y}$ can be inaccurate in either prediction or manual annotation) and formulation ({\em e.g.}, perspective projection does not model camera distortion), Eqn~\eqref{Eqn:AffineTransformation} may not hold perfectly. In practice, we assume the recovered 3D coordinates to follow an isotropic Gaussian distribution, and so maximizing its likelihood gives the following log-likelihood loss:
\begin{equation}
\label{Eqn:ReconstructionLoss}
{\mathcal{L}\!\left(\mathbf{p},\mathbf{z}\mid\mathbf{y}\right)}={\left\|\left[\mathbf{y}|\mathbf{1}\right]^\top\cdot\hat{\mathbf{S}}-\mathbf{K}\cdot\left[\mathbf{R}|\mathbf{T}\right]\cdot\left[\mathbf{z}|\mathbf{1}\right]^\top\right\|_2^2}.
\end{equation}


\section{Experiments}
\label{Experiments}

\begin{figure}
\centering
\includegraphics[width=\linewidth]{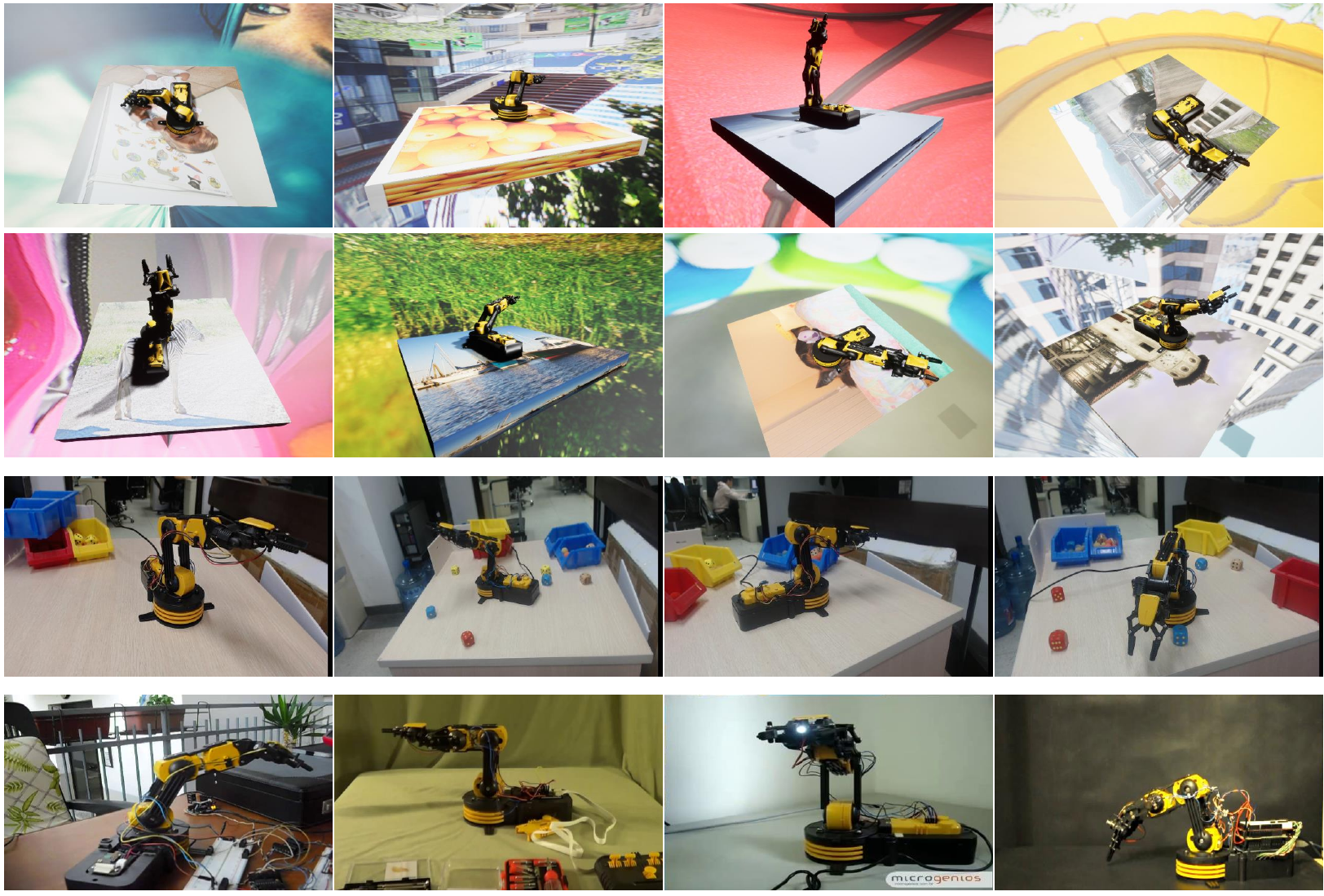}
\caption{Example images of three datasets used in this paper. From top to bottom: synthetic images (top two rows), lab images and YouTube images. Please zoom in to see details.}
\label{Fig:DataExamples}
\vspace{-0.4cm}
\end{figure}

\subsection{Dataset and Settings}
\label{Experiments:DatasetSettings}

We generated $5\rm{,}000$ synthetic images with randomized camera parameters, lighting conditions, arm poses and background augmentation (see Section~\ref{Approach:Details}). Among them, $4\rm{,}500$ are used for training and the remaining $500$ for validation. This dataset, later referred to as the {\bf virtual dataset}, is used to verify the 2D keypoint detection model, {\em e.g.}, a stacked hourglass network, works well.

In the real environments, we collected and manually annotated two sets of data. The first one is named the {\bf lab dataset}, which contains more than $20\rm{,}000$ frames captured by a $720$P Webcam. We manually chose $428$ key frames and annotated them. For this purpose, we rendered the 3D model of the arm in the virtual environment and adjusted it to the same pose of the real arm so that arms in these two images exactly overlap with each other -- in this way we obtained the ground-truth arm pose, as well as the camera intrinsic and extrinsic (obtained by a checkerboard placed alongside the robotic arm at the beginning of each video) parameters. We deliberately put distractors, e.g. colorful boxes, dices and balls, to make the dataset more difficult and thus evaluate the generalization ability of our models. The frames used for fine-tuning and for testing come from different videos.

The second part of real-world image data, the {\bf YouTube dataset}, is crawled from YouTube, which contains $109$ videos with the OWI-535 arm. This is a largely diversified collection, in which the arm may even be modded, {\em i.e.}, the geometric constraints may not hold perfectly. We sampled $275$ frames and manually annotated the visibility as well as position for each 2D keypoint. Note that, without camera parameters, we cannot annotate the accurate pose of the arm. This dataset is never included in training, but used to observe the behavior of domain adaptation.

Sample images of three datasets are shown in Figure~\ref{Fig:DataExamples}.

\subsection{Pose Estimation}
\label{Experiments:PoseEstimation}

\subsubsection{Detecting 2D Keypoints}
\label{Experiments:PoseEstimation:Keypoints}

We first evaluate 2D keypoint detection, and make use of a popular metric named PCK@0.2~\cite{yang2013articulated} to evaluate the accuracy. For this purpose, we train a $2$-stack hourglass network from scratch for $30$ epochs in the {\bf virtual} dataset. Standard data augmentation techniques are applied, including random translation, rotation, scaling, color shifting and flipping. On top of this model, we consider several approaches to achieve domain transfer. One is to train an explicit model which transfers virtual data to fake real data on which we train a new model. In practice, we apply a popular generative model named CycleGAN~\cite{zhu2017unpaired}. We trained the CycleGAN network with synthetic image as the source domain and lab image as the target domain for 100 epochs. Other domain adaptation methods, {\em i.e.}, ADDA \cite{tzeng2017adversarial}  and  its  follow-up  work  CyCADA \cite{hoffman2018cycada} are also applied and compared with our approach described in Section~\ref{Approach:PoseEstimation}. We mix the synthetic and real images with a ratio of $6:4$ and use the same hyper-parameters as for the baseline. Background clutters are added to the lab images in an online manner to facilitate variability (see Section~\ref{Approach:Details}).

\begin{table}
\centering
\setlength{\tabcolsep}{2pt}
\resizebox{0.95\linewidth}{!}{
\begin{tabular}{|c||c|c|c|c|}
\hline
Model & Virtual & Lab & YouTube & YouTube-vis\\
\hline\hline
Synthetic  & \textbf{99.95} & 95.66          & 80.05          & 81.61          \\ \hline
CycleGAN   & 99.86          & 97.84          & 75.26          & 76.98          \\ \hline
ADDA       & 99.89          & 96.14          & 79.19          & 80.04          \\ \hline
CyCADA     & 99.84          & 98.09          & 73.47          & 74.37          \\ \hline
Our Method & 99.63          & \textbf{99.55} & \textbf{87.01} & \textbf{88.89} \\ \hline

\end{tabular}
}


\caption{2D keypoint detection accuracy (PCK@0.2, \%) on three datasets. Models are tested on YouTube dataset when considering all keypoints and considering only the visible ones. }
\label{tab:result_real_2d}
\vspace{-0.2cm}
\end{table}

Results are summarized in Table~\ref{tab:result_real_2d}. The baseline model works almost perfectly on \textbf{virtual} data, which reports a PCK@$0.2$ accuracy of $99.95\%$. However, this number drops significantly to $95.66\%$ in lab data, and even dramatically to $80.05\%$ in \textbf{YouTube} data, demonstrating the existence of domain gaps. These gaps are largely shrunk after domain adaptation algorithms are applied. Training with images generated by CycleGAN, we found that the model works better in its target domain, i.e. the \textbf{lab} dataset by a margin of $2.18\%$. However, this model failed to generalize to \textbf{YouTube} dataset, as the accuracy is even lower than the baseline model. The cases are similar for ADDA and CyCADA, which gains $0.48\%$ and $2.43\%$ improvement on the \textbf{lab} dataset respectively, but both of the approaches do not generalize well to the \textbf{YouTube} dataset. Our approach, on the other hand, achieves much higher accuracy, with a PCK@$0.2$ score of $99.55\%$ in the {\bf lab} data, and $87.01\%$ in the {\bf YouTube}, boosting the baseline performance by $6.96\%$. In the subset of visible {\bf YouTube} keypoints, the improvement is even higher ($7.28\%$). In addition, the refined model only produces a slightly worse PCK@$0.2$ accuracy ($99.63\%$ vs. $99.95\%$) on \textbf{virtual} data, implying that balance is achieved between ``fitting on virtual data'' and ``transferring to real data''. 

The results reveal that performance of explicit domain adaptation manners, {\em i.e.} CycleGAN, ADDA and CyCADA can be limited in several aspect. For instance, compared with our 3D geometric based domain adaptation method, although the models trained with these explicit domain adaptation methods fit to the target domain, it has a poor performance on unseen data. Moreover, we fail to train a CycleGAN model with \textbf{YouTube} dataset as the target domain, because the distribution of data in YouTube is too diverse and such transformation is hard to learn.


\subsubsection{Estimating the 3D Pose}
\label{Experiments:PoseEstimation:Pose}


\begin{table}
\centering

\setlength{\tabcolsep}{2pt}
\resizebox{\linewidth}{!}{
\begin{tabular}{|c||ccccc|cc|}
\hline
 \multirow{2}{*}{Model} & \multicolumn{5}{c|}{Motor} & \multicolumn{2}{c|}{Camera}\\
  & Rotation & Base & Elbow & Wrist& Average & Rotation & Location  \\
\hline\hline
Synthetic & 7.41  & 6.20 & 7.15  & 7.74   &  7.13  &  6.29  & 7.58 \\
\hline
Refined & \textbf{4.40} & \textbf{3.29} & \textbf{5.35}   &  \textbf{6.03}   &  \textbf{4.81}   &  \textbf{5.29}  & \textbf{6.73}\\
\hline
\end{tabular}
}
\caption{3D pose estimation errors (degrees) and camera parameter prediction errors (degrees and centimeters) in the {\bf lab} dataset.}
\label{tab:result_real_3d}
\vspace{-0.2cm}
\end{table}

\begin{figure}
\centering
\includegraphics[width=\linewidth]{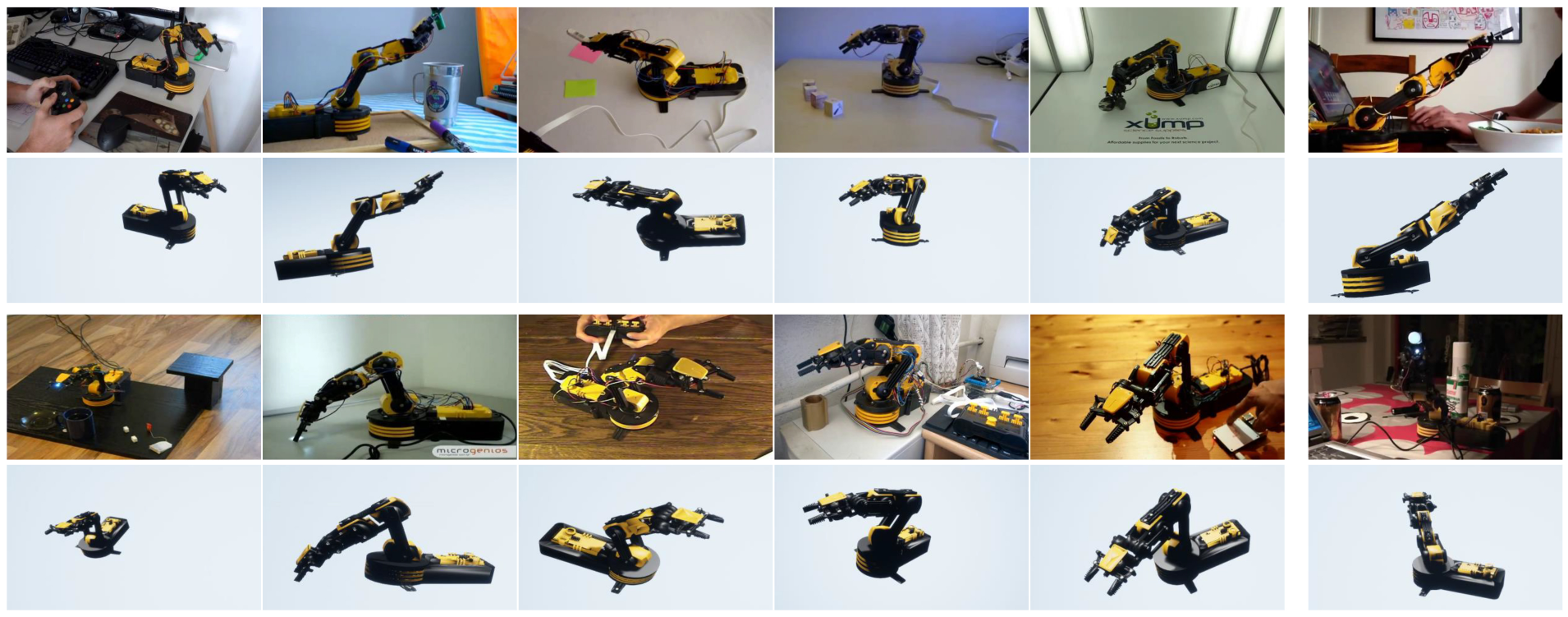}
\caption{Qualitative results from our YouTube dataset. The challenges include occlusion, user modification, lighting, etc. We show synthetic images generated using the camera parameters and pose estimated from the single input image. Both success cases (left five columns) and failure cases (rightmost column) are shown.}
\label{fig:3d_youtube}
\vspace{-0.4cm}
\end{figure}

We first test the performance of 3D pose estimation on the \textbf{virtual} dataset. We use the model trained only on synthetic images since it has the best 2D prediction accuracy on synthetic data. The experiment was conducted on $500$ synthetic images. The angular error for four joints are $2.67\degree$, $2.80\degree$, $2.76\degree$, $3.31\degree$, with an average of $2.89\degree$. The error of camera parameters is $2.25\degree$ for rotation and $2.59$cm for location. 

We also test the 3D pose estimation performance of our model on real images, which is the basis for completing tasks. The quantitative 3D pose estimation result is only reported for the \textbf{lab} dataset, since getting 3D annotation for \textbf{YouTube} data is difficult. We estimate the camera intrinsic parameters by using camera calibration with checkerboard. Results are shown in Table~\ref{tab:result_real_3d}. The results reveal that our refined model outperforms the synthetic only model by $\ang{2.32}$ on average angular error. The qualitative result on \textbf{YouTube} dataset are shown in Figure \ref{fig:3d_youtube}. Since the camera intrinsic parameters are unknown for YouTube videos, we use the weak perspective model during reconstruction. Heavy occlusion, user modification and extreme lighting make the 3D pose estimation hard in some cases. We select typical samples for success and failure cases.

\subsubsection{Ablation Study: Domain Adaptation Options}

As described in Section~\ref{Approach:Details}, when training the refined model, two strategies are applied on the real images: random background augmentation and joint keypoint detection and pose estimation. To evaluate the contribution of these two strategies to the improvement on accuracy, we did an ablation study. Results are shown in Table \ref{tab:ablation_study}.

We compare the performance of 5 models: 1) baseline model, trained on $4\rm{,}500$ synthetic images; 2) - 5) models trained with/without joint estimation and with/without background augmentation. Note that if a model is trained without joint estimation, we directly take the argmax on the predicted heatmap as annotations for training.  
We see that both strategies contribute to the overall performance improvement. Our model performs best when combining both strategies. 

\begin{table}
\centering
\begin{tabular}{|l l||c|c|c|c|}
\hline
\multicolumn{2}{|c||}{Model} &  Lab & YouTube & YouTube-vis\\
\hline\hline
\multicolumn{2}{|c||}{Synthetic Only} & 95.66 & 80.05 & 81.61 \\
\hline
\hspace{0pt} BG \xmark & 3D \xmark & 94.52 & 80.24 & 82.22\\
\hline
\hspace{0pt} BG \cmark & 3D \xmark & 98.72 & 84.04 & 87.11 \\
\hline
\hspace{0pt} BG \xmark & 3D \cmark & 97.31 & 86.52 & 88.27 \\
\hline
\hspace{0pt} BG \cmark & 3D \cmark & \textbf{99.55} & \textbf{87.01} & \textbf{88.89}\\
\hline
\end{tabular}
\caption{2D keypoint detection accuracy (PCK@0.2, \%) under ablation study. `3D' stands for joint keypoint detection and 3D pose estimation, and `BG' for random background augmentation.}
\label{tab:ablation_study}
\vspace{-0.2cm}
\end{table}

\subsubsection{Ablation Study: Number of Training Images}

With the help of domain randomization, we can generate an unlimited number of synthetic images with high abundance in their appearance. On the other hand, the performance of deep models tends to saturate as the number of training data increases. Therefore, it is necessary to balance between performance and data-efficiency.

Results are shown in Table~\ref{tab:num_images}. As the number of training images increases from $2\rm{,}500$ to $5\rm{,}000$, the accuracy significantly increases (by $3.28\%$ and $4.81\%$ on \textbf{lab} and \textbf{YouTube} dataset, respectively). When the number of training images continue to increase to $20\rm{,}000$, the accuracy only increases by a small margin ($1.49\%$ and $0.22\%$ on \textbf{lab} and \textbf{YouTube} dataset, respectively). Therefore, $5\rm{,}000$ synthetic images is a nice balance between accuracy and efficiency.

\begin{table}[]
\centering
{
\begin{tabular}{|c||c|c|c|}
\hline
\# Images               & Lab            & YouTube        & YouTube-vis    \\ \hline \hline
2,500  & 92.38 & 75.24 & 77.33 \\ \hline
5,000  & 95.66 & 80.05 & 81.61 \\ \hline
10,000 & 96.13 & 80.71 & 81.71 \\ \hline
20,000 & 97.25 & 81.11 & 81.83 \\ \hline
\end{tabular}}
\caption{2D keypoint detection accuracy (PCK@0.2, \%) with respect to the number of training images.}
\vspace{-0.2cm}
\label{tab:num_images}
\end{table}

\subsection{Controlling the Arm with Vision}

We implement a complete control system purely based on vision, as described in Section~\ref{Approach:MotionPlanning}. It takes a video stream as input, estimates the arm pose, then plans the motion and sends control signal to the motors. 

\begin{figure}
\centering
\includegraphics[width=\linewidth]{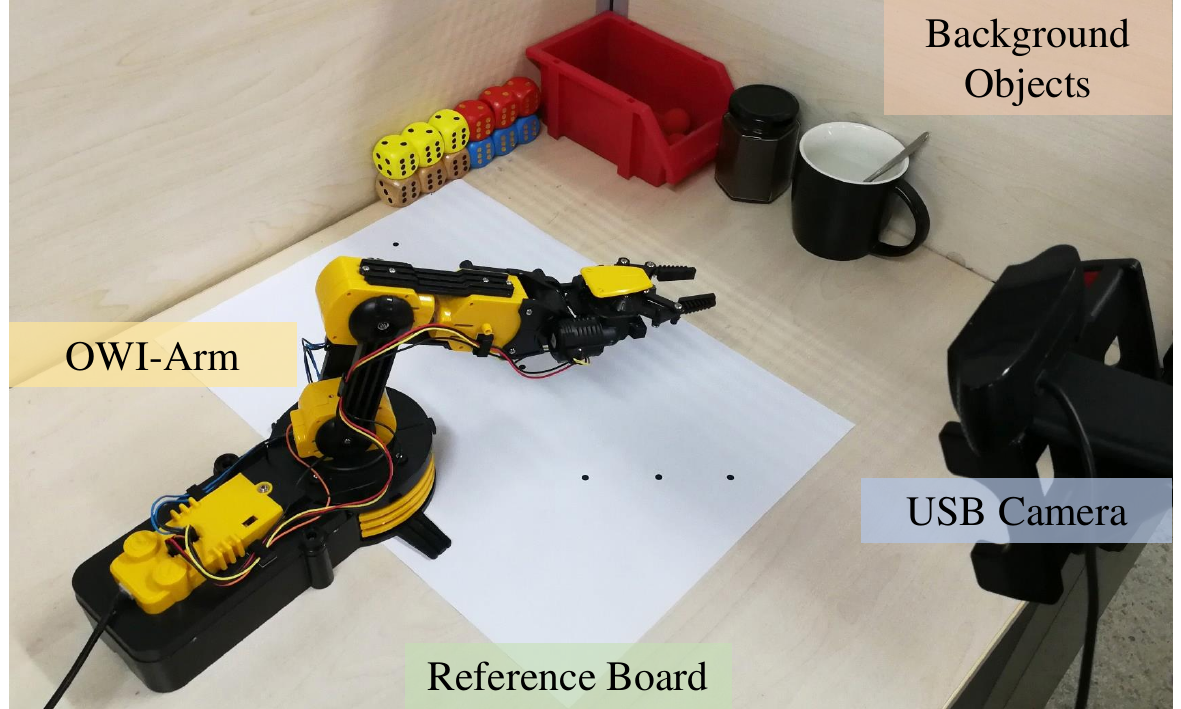}
\caption{A snapshot of the real-world experiment setup. Locations of the goals are printed on the reference board and are used as reference when measuring the error. We scatter background objects randomly during testing.}
\label{fig:real-experiment}
\vspace{-0.4cm}
\end{figure}

This system is verified with a task, reaching a target point.
The goal is controlling the arm to make the arm tip reach right above a specified point on the table without collision. Each attempt is considered successful if the horizontal distance between arm tip and the target is within 3cm. The system is tested at 6 different camera views. For each view, the arm needs to reach 9 target points. The target points and camera views are selected to cover a variety of cases. We also place distractors to challenge our vision module. A snapshot of our experiment setup can be seen in Figure~\ref{fig:real-experiment}.

We report human performance on the same task. Human is asked to watch the video stream from a screen and control the arm with a game pad. This setup ensures human and our system accepts the same vision input in comparison. In addition, we allow human to directly look at the arm and move freely to observe the arm when doing the task. The performance for both setups are reported.

Our control system can achieve comparable performance with human in this task. The result is reported in Table~\ref{tab:task_real}. Human can perform much better if directly looking at the arm. This is because human can constantly move his head to pick the best view for current state. Potentially, we could use action vision, or multi-camera system, to mimic this ability and further improve the system, which are interesting future work but beyond the scope of this work. It is worth noticing our system can run real-time and finish the task faster than human.

Built on this control module, we show that our system can move a stack of dices into a box separately. Please see {\sf https://youtu.be/8hZjdqDrYas} for video demonstration. To simplify deployment, we directly feed the ground-truth locations of the dices and the box into the system and apply the same controller as the reaching task. The successful rate of this task is not high, because it requires highly accurate pose control in both horizontal and vertical directions. We also provide some interesting failure cases in the video. Failure cases are caused by several reasons, {\em e.g.}, self occlusion or rarely seen configurations. Also, the controller sometimes fails, {\em e.g.}, if the arm is too far from the camera, a long-distance movement only causes minor visual difference. At the current point, this demo reveals the potential of our vision-based control system, as well as advocates more advanced vision algorithms to be designed to improve its performance.

\begin{table}
\centering
\begin{tabular}{|c|c||c|c|c|}
\hline
\multirow{2}*{Agent} & Input & Distance & Success & Average \\
 & Type & Error (cm) & Rate & Time (s) \\
\hline\hline
Human & Direct & 0.65 & 100.0\% & 29.8\\
\hline\hline
Human & Camera & 2.67 & \textbf{66.7\%} & 38.8 \\
\hline
Ours & Camera & \textbf{2.66} & 59.3\% & \textbf{21.2} \\
\hline
\end{tabular}
\caption{Quantitative result for completing the reaching task. Our system achieves comparable performance with human when the same input signal is given.}
\label{tab:task_real}
\vspace{-0.2cm}
\end{table}

\section{Conclusions}
\label{Conclusions}

In this paper, we built a system, which is purely based on vision inputs, to control a low-cost, sensor-free robotic arm to accomplish tasks. We used a semi-supervised algorithm which integrates labeled synthetic as well as unlabeled real data to train the pose estimation module. Geometric constraints of multi-rigid-body system (the robotic arm in this case) was utilized for domain adaptation. Our approach, with merely a 3D model being required, has the potential to be applied to other multi-rigid-body systems.

To facilitate reproducible research, we created a virtual environment to generate synthetic data, and also collected two real-world datasets from our lab and YouTube videos, respectively, all of which can be used as benchmarks to evaluate 2D keypoint detection and/or 3D pose estimation algorithms. In addition, the low cost of our system enables vision researchers to study robotic tasks, {\em e.g.}, reinforcement learning, imitation learning, active vision, {\em etc.}, without large economic expenses. This system also has the potential to be used for high-school and college educational purposes.

Beyond our work, many interesting future directions can be explored. For example, we can train perception and controller modules in a joint, end-to-end manner~\cite{levine2016end}\cite{jang2017end}, or incorporate other vision components, such as object detection and 6D pose estimation, to enhance the ability of the arm so that more complex tasks can be accomplished.

\vspace{-0.3cm}
\section*{Acknowledgements}
\vspace{-0.2cm}
This work is supported by IARPA via DOI/IBC contract No. D17PC00342, NSFC-61625201, 61527804, Qualcomm University Research Grant and CCF-Tencent Open Fund. The authors would like to thank Vincent Yan, Kaiyue Wu, Andrew Hundt, Prof. Yair Amir and Prof. Gregory Hager for helpful discussions.






{\small
\bibliographystyle{ieee}
\bibliography{egbib}

\begin{thebibliography}{10}\itemsep=-1pt

\bibitem{bogo2016keep}
Federica Bogo, Angjoo Kanazawa, Christoph Lassner, Peter Gehler, Javier Romero,
  and Michael~J Black.
\newblock Keep it smpl: Automatic estimation of 3d human pose and shape from a
  single image.
\newblock In {\em European Conference on Computer Vision}, 2016.

\bibitem{bousmalis2018using}
Konstantinos Bousmalis, Alex Irpan, Paul Wohlhart, Yunfei Bai, Matthew Kelcey,
  Mrinal Kalakrishnan, Laura Downs, Julian Ibarz, Peter Pastor, Kurt Konolige,
  et~al.
\newblock Using simulation and domain adaptation to improve efficiency of deep
  robotic grasping.
\newblock In {\em International Conference on Robotics and Automation}, 2018.

\bibitem{butler2012naturalistic}
Daniel~J Butler, Jonas Wulff, Garrett~B Stanley, and Michael~J Black.
\newblock A naturalistic open source movie for optical flow evaluation.
\newblock In {\em European Conference on Computer Vision}, 2012.

\bibitem{csurka2017domain}
Gabriela Csurka.
\newblock Domain adaptation for visual applications: A comprehensive survey.
\newblock {\em arXiv preprint arXiv:1702.05374}, 2017.

\bibitem{deisenroth2012learning}
Marc~Peter Deisenroth.
\newblock Learning to control a low-cost manipulator using data-efficient
  reinforcement learning.
\newblock {\em Robotics: Science and Systems}, 2012.

\bibitem{gaidon2016virtual}
Adrien Gaidon, Qiao Wang, Yohann Cabon, and Eleonora Vig.
\newblock Virtual worlds as proxy for multi-object tracking analysis.
\newblock In {\em Conference on Computer Vision and Pattern Recognition}, 2016.

\bibitem{gao2017exploiting}
Yuan Gao and Alan~L Yuille.
\newblock Exploiting symmetry and/or manhattan properties for 3d object
  structure estimation from single and multiple images.
\newblock In {\em Conference on Computer Vision and Pattern Recognition}, 2017.

\bibitem{geiger2013vision}
Andreas Geiger, Philip Lenz, Christoph Stiller, and Raquel Urtasun.
\newblock Vision meets robotics: The kitti dataset.
\newblock {\em The International Journal of Robotics Research},
  32(11):1231--1237, 2013.

\bibitem{hoffman2018cycada}
Judy Hoffman, Eric Tzeng, Taesung Park, Jun{-}Yan Zhu, Phillip Isola, Kate
  Saenko, Alexei~A. Efros, and Trevor Darrell.
\newblock Cycada: Cycle-consistent adversarial domain adaptation.
\newblock In {\em International Conference on Machine Learning}, 2018.

\bibitem{hoffman2016fcns}
Judy Hoffman, Dequan Wang, Fisher Yu, and Trevor Darrell.
\newblock Fcns in the wild: Pixel-level adversarial and constraint-based
  adaptation.
\newblock {\em arXiv preprint arXiv:1612.02649}, 2016.

\bibitem{hong2018virtual}
Zhang{-}Wei Hong, Yu{-}Ming Chen, Hsuan{-}Kung Yang, Shih{-}Yang Su, Tzu{-}Yun
  Shann, Yi{-}Hsiang Chang, Brian~Hsi{-}Lin Ho, Chih{-}Chieh Tu, Tsu{-}Ching
  Hsiao, Hsin{-}Wei Hsiao, Sih{-}Pin Lai, Yueh{-}Chuan Chang, and Chun{-}Yi
  Lee.
\newblock Virtual-to-real: Learning to control in visual semantic segmentation.
\newblock In {\em International Joint Conference on Artificial Intelligence},
  2018.

\bibitem{hundt2018training}
Andrew Hundt, Varun Jain, Chris Paxton, and Gregory~D Hager.
\newblock Training frankenstein's creature to stack: Hypertree architecture
  search.
\newblock {\em arXiv preprint arXiv:1810.11714}, 2018.

\bibitem{jang2017end}
Eric Jang, Sudheendra Vijaynarasimhan, Peter Pastor, Julian Ibarz, and Sergey
  Levine.
\newblock End-to-end learning of semantic grasping.
\newblock {\em arXiv preprint arXiv:1707.01932}, 2017.

\bibitem{johnson2017clevr}
Justin Johnson, Bharath Hariharan, Laurens van~der Maaten, Li Fei-Fei,
  C~Lawrence Zitnick, and Ross Girshick.
\newblock Clevr: A diagnostic dataset for compositional language and elementary
  visual reasoning.
\newblock In {\em Conference on Computer Vision and Pattern Recognition}, 2017.

\bibitem{kahn2017uncertainty}
Gregory Kahn, Adam Villaflor, Vitchyr Pong, Pieter Abbeel, and Sergey Levine.
\newblock Uncertainty-aware reinforcement learning for collision avoidance.
\newblock {\em arXiv preprint arXiv:1702.01182}, 2017.

\bibitem{koenig2004design}
Nathan~P Koenig and Andrew Howard.
\newblock Design and use paradigms for gazebo, an open-source multi-robot
  simulator.
\newblock In {\em International Conference on Intelligent Robots and Systems},
  2004.

\bibitem{lecun2015deep}
Yann LeCun, Yoshua Bengio, and Geoffrey Hinton.
\newblock Deep learning.
\newblock {\em Nature}, 521(7553):436, 2015.

\bibitem{levine2016end}
Sergey Levine, Chelsea Finn, Trevor Darrell, and Pieter Abbeel.
\newblock End-to-end training of deep visuomotor policies.
\newblock {\em The Journal of Machine Learning Research}, 17(1):1334--1373,
  2016.

\bibitem{levine2018learning}
Sergey Levine, Peter Pastor, Alex Krizhevsky, Julian Ibarz, and Deirdre
  Quillen.
\newblock Learning hand-eye coordination for robotic grasping with deep
  learning and large-scale data collection.
\newblock {\em The International Journal of Robotics Research},
  37(4-5):421--436, 2018.

\bibitem{li2017deep}
Chi Li, M.~Zeeshan Zia, Quoc{-}Huy Tran, Xiang Yu, Gregory~D. Hager, and
  Manmohan Chandraker.
\newblock Deep supervision with shape concepts for occlusion-aware 3d object
  parsing.
\newblock In {\em Conference on Computer Vision and Pattern Recognition}, 2017.

\bibitem{ddpg}
Timothy~P Lillicrap, Jonathan~J Hunt, Alexander Pritzel, Nicolas Heess, Tom
  Erez, Yuval Tassa, David Silver, and Daan Wierstra.
\newblock Continuous control with deep reinforcement learning.
\newblock In {\em International Conference for Learning Representations}, 2016.

\bibitem{lin2014microsoft}
Tsung-Yi Lin, Michael Maire, Serge Belongie, James Hays, Pietro Perona, Deva
  Ramanan, Piotr Doll{\'a}r, and C~Lawrence Zitnick.
\newblock Microsoft coco: Common objects in context.
\newblock In {\em European conference on computer vision}, 2014.

\bibitem{luo2018end}
Wenhan Luo, Peng Sun, Fangwei Zhong, Wei Liu, Tong Zhang, and Yizhou Wang.
\newblock End-to-end active object tracking via reinforcement learning.
\newblock In {\em International Conference on Machine Learning}, pages
  3286--3295, 2018.

\bibitem{luo2018journal}
Wenhan Luo, Peng Sun, Fangwei Zhong, Wei Liu, Tong Zhang, and Yizhou Wang.
\newblock End-to-end active object tracking and its real-world deployment via
  reinforcement learning.
\newblock {\em IEEE Transactions on Pattern Analysis and Machine Intelligence},
  2019.

\bibitem{mahmood2018unsupervised}
Faisal Mahmood, Richard Chen, and Nicholas~J Durr.
\newblock Unsupervised reverse domain adaptation for synthetic medical images
  via adversarial training.
\newblock {\em IEEE Transactions on Medical Imaging}, 2018.

\bibitem{newell2016stacked}
Alejandro Newell, Kaiyu Yang, and Jia Deng.
\newblock Stacked hourglass networks for human pose estimation.
\newblock In {\em European Conference on Computer Vision}, 2016.

\bibitem{omran2018neural}
Mohamed Omran, Christoph Lassner, Gerard Pons-Moll, Peter Gehler, and Bernt
  Schiele.
\newblock Neural body fitting: Unifying deep learning and model based human
  pose and shape estimation.
\newblock In {\em International Conference on 3D Vision}, 2018.

\bibitem{pavlakos2018learning}
Georgios Pavlakos, Luyang Zhu, Xiaowei Zhou, and Kostas Daniilidis.
\newblock Learning to estimate 3d human pose and shape from a single color
  image.
\newblock In {\em Conference on Computer Vision and Pattern Recognition}, 2018.

\bibitem{pinto2016supersizing}
Lerrel Pinto and Abhinav Gupta.
\newblock Supersizing self-supervision: Learning to grasp from 50k tries and
  700 robot hours.
\newblock In {\em International Conference on Robotics and Automation}, 2016.

\bibitem{qiu2017unrealcv}
Weichao Qiu, Fangwei Zhong, Yi Zhang, Siyuan Qiao, Zihao Xiao, Tae~Soo Kim,
  Yizhou Wang, and Alan Yuille.
\newblock Unrealcv: Virtual worlds for computer vision.
\newblock In {\em Proceedings of the 25th ACM International Conference on
  Multimedia}, 2017.

\bibitem{rahmatizadeh2018vision}
Rouhollah Rahmatizadeh, Pooya Abolghasemi, Ladislau B{\"o}l{\"o}ni, and Sergey
  Levine.
\newblock Vision-based multi-task manipulation for inexpensive robots using
  end-to-end learning from demonstration.
\newblock In {\em International Conference on Robotics and Automation}, 2018.

\bibitem{rusu2016sim}
Andrei~A Rusu, Matej Vecerik, Thomas Roth{\"o}rl, Nicolas Heess, Razvan
  Pascanu, and Raia Hadsell.
\newblock Sim-to-real robot learning from pixels with progressive nets.
\newblock {\em arXiv preprint arXiv:1610.04286}, 2016.

\bibitem{shrivastava2017learning}
Ashish Shrivastava, Tomas Pfister, Oncel Tuzel, Joshua Susskind, Wenda Wang,
  and Russell Webb.
\newblock Learning from simulated and unsupervised images through adversarial
  training.
\newblock In {\em Conference on Computer Vision and Pattern Recognition}, 2017.

\bibitem{su2015render}
Hao Su, Charles~R Qi, Yangyan Li, and Leonidas~J Guibas.
\newblock Render for cnn: Viewpoint estimation in images using cnns trained
  with rendered 3d model views.
\newblock In {\em International Conference on Computer Vision}, 2015.

\bibitem{sundermeyer2018implicit}
Martin Sundermeyer, Zoltan-Csaba Marton, Maximilian Durner, Manuel Brucker, and
  Rudolph Triebel.
\newblock Implicit 3d orientation learning for 6d object detection from rgb
  images.
\newblock In {\em European Conference on Computer Vision}, 2018.

\bibitem{tobin2017domain}
Josh Tobin, Rachel Fong, Alex Ray, Jonas Schneider, Wojciech Zaremba, and
  Pieter Abbeel.
\newblock Domain randomization for transferring deep neural networks from
  simulation to the real world.
\newblock In {\em International Conference on Intelligent Robots and Systems},
  2017.

\bibitem{todorov2012mujoco}
Emanuel Todorov, Tom Erez, and Yuval Tassa.
\newblock Mujoco: A physics engine for model-based control.
\newblock In {\em International Conference on Intelligent Robots and Systems},
  2012.

\bibitem{tremblay2018training}
Jonathan Tremblay, Aayush Prakash, David Acuna, Mark Brophy, Varun Jampani, Cem
  Anil, Thang To, Eric Cameracci, Shaad Boochoon, and Stan Birchfield.
\newblock Training deep networks with synthetic data: Bridging the reality gap
  by domain randomization.
\newblock {\em arXiv preprint arXiv:1804.06516}, 2018.

\bibitem{tzeng2015adapting}
Eric Tzeng, Coline Devin, Judy Hoffman, Chelsea Finn, Pieter Abbeel, Sergey
  Levine, Kate Saenko, and Trevor Darrell.
\newblock Adapting deep visuomotor representations with weak pairwise
  constraints.
\newblock {\em arXiv preprint arXiv:1511.07111}, 2015.

\bibitem{tzeng2017adversarial}
Eric Tzeng, Judy Hoffman, Kate Saenko, and Trevor Darrell.
\newblock Adversarial discriminative domain adaptation.
\newblock In {\em Conference on Computer Vision and Pattern Recognition}, 2017.

\bibitem{varol2017learning}
G{\"u}l Varol, Javier Romero, Xavier Martin, Naureen Mahmood, Michael~J Black,
  Ivan Laptev, and Cordelia Schmid.
\newblock Learning from synthetic humans.
\newblock In {\em Conference on Computer Vision and Pattern Recognition}, 2017.

\bibitem{yang2013articulated}
Yi Yang and Deva Ramanan.
\newblock Articulated human detection with flexible mixtures of parts.
\newblock {\em IEEE transactions on pattern analysis and machine intelligence},
  35(12):2878--2890, 2013.

\bibitem{yu2018bdd100k}
Fisher Yu, Wenqi Xian, Yingying Chen, Fangchen Liu, Mike Liao, Vashisht
  Madhavan, and Trevor Darrell.
\newblock Bdd100k: A diverse driving video database with scalable annotation
  tooling.
\newblock {\em arXiv preprint arXiv:1805.04687}, 2018.

\bibitem{zhong2018advat}
Fangwei Zhong, Peng Sun, Wenhan Luo, Tingyun Yan, and Yizhou Wang.
\newblock {AD}-{VAT}: An asymmetric dueling mechanism for learning visual
  active tracking.
\newblock In {\em International Conference on Learning Representations}, 2019.

\bibitem{zhu2017unpaired}
Jun{-}Yan Zhu, Taesung Park, Phillip Isola, and Alexei~A. Efros.
\newblock Unpaired image-to-image translation using cycle-consistent
  adversarial networks.
\newblock In {\em International Conference on Computer Vision}, 2017.

\bibitem{zhu2006semi}
Xiaojin Zhu.
\newblock Semi-supervised learning literature survey.
\newblock {\em Computer Science, University of Wisconsin-Madison}, 2006.

\bibitem{zuffi2018lions}
Silvia Zuffi, Angjoo Kanazawa, and Michael~J Black.
\newblock Lions and tigers and bears: Capturing non-rigid, 3d, articulated
  shape from images.
\newblock In {\em Conference on Computer Vision and Pattern Recognition}, 2018.

\end{thebibliography}
}

\end{document}